\documentclass[11pt]{article}
\usepackage{authblk}
\usepackage[final]{acl}
\usepackage{times,latexsym}
\usepackage{url}
\usepackage[T1]{fontenc}
\usepackage{todonotes}
\usepackage{xspace,mfirstuc,tabulary}
\usepackage{times}
\usepackage{latexsym}
\usepackage{subcaption,graphicx}
\usepackage{tcolorbox}
\usepackage{url}
\usepackage{amsmath,amssymb}
\usepackage{mathtools}
\usepackage{dirtytalk}
\usepackage{cleveref}
\usepackage{alphalph}
\usepackage{multirow}
\usepackage{multicol}
\usepackage{caption}
\usepackage{comment}
\usepackage{adjustbox}
\usepackage{xcolor}
\usepackage{pifont}
\usepackage{cleveref}
\usepackage{todonotes}

\usepackage[utf8]{inputenc}
\usepackage{microtype}
\usepackage{enumitem,kantlipsum}

\crefformat{section}{\S#2#1#3} 
\crefformat{subsection}{\S#2#1#3}

\label{firstpage}

\title{On the Origins of Bias in NLP through the Lens of the Jim Code}

\author[1]{Fatma Elsafoury}
\author[2]{Gavin Abercrombie}
\affil[1]{\small{School of Physics, Engineering, and Computing, the University of The West of Scotland, UK}}
\affil[2]{\small{HUW NLP lab, Heriot-Watt University, UK}}

\begin{document}
\maketitle
\begin{abstract}
In this paper, we trace the biases in current natural language processing (NLP) models back to 
their origins in racism, sexism, and homophobia over
the last 500 years. We review literature from critical race theory, gender studies, data ethics, and digital humanities studies, and summarize the origins of bias in NLP models from these social science perspective. We show how the causes of the biases in the NLP pipeline
are rooted in social issues. Finally, we 
argue that the only way to fix the bias and unfairness in NLP is by addressing the social problems that caused them in the first place and by incorporating social sciences and social scientists in efforts to mitigate bias in NLP models.
We provide actionable recommendations for the NLP research community to do so.
\end{abstract}
\section{Introduction}
In 
\textit{Race After Technology}, 
\citet{benjamin2019race}
coins the term \say{The New Jim Code}, which she describes 
as :
\begin{quote}
\textit{\say{The employment of new technologies that reflect and reproduce existing inequities but that are promoted and perceived as more objective or progressive than discriminatory systems of a previous era.}}
\end{quote}
While
the Jim Code is a spin on, \say{Jim Crow},
a derogatory epithet for African-Americans, the same concept can be generalized to the bias and unfairness in artificial intelligence (AI) systems against all marginalised groups. It is crucial to study bias and fairness in machine learning (ML) and natural language processing (NLP) models to understand how existing social biases and stereotyping are being encoded in the data used to train
them,  
as well as to compare
(1) the fairness of
the decisions made by NLP models due to biases in the datasets, with (2) biased choices made by the developers of those models as a result of unintended bias or to maximize profit. Studying bias and unfairness in NLP models is one way to pierce a hole in the black box and shed a little light on the limitations of widely used models. However, it 
is not 
possible to 
understand the roots of algorithmic bias, without incorporating relevant studies from social sciences, critical race theory, gender studies, LGBTQ studies, and digital humanities studies, as 
recommended
by~\citet{benjamin2019race}.

In this paper, 
we study the various origins of bias in NLP models from two perspectives: 
(1)~the \emph{NLP pipeline perspective}, where we review the sources of bias in models from the NLP literature; and 
(2)~the \emph{Jim Code perspective}, in which we review the origins of bias from the literature on social science, critical race theory, gender, LGBTQ, and digital studies. 
We argue that, in fact, the sources of bias found in the NLP pipeline 
are rooted in 
those uncovered in the
social sciences. 
Then, we discuss how the lack of inclusion of social sciences in attempts at eliminating social issues like bias in NLP models
has
resulted in problematic quantitative measures of bias \cite{Blodgett-etal-2021-norweigan-salmon} and superficial mitigation techniques \cite{gonen-goldberg-2019-lipstick}. Finally, we propose recommendations to NLP researchers to mitigate biases and improve the fairness of the models that they develop by addressing their underlying social causes.  

\section{Background: History of discrimination} 
In Western societies, the biases and inequalities towards marginalised groups based on ethnicity, sex, class, religion, sexual orientation, age, or disability that we see today are direct results of centuries of racism, sexism, and homophobia, as has been discussed by many scholars. 

In 
\textit{The Myth of Race: The Troubling Persistence of an Unscientific Idea}, \citet{10.2307/j.ctt9qdt73.5} 
reviews the history of 500 years of \textbf{racism} in Western Europe to answer the question of why the 
invalid concept
of race still prevails. 
He
argues that 
the ideology of race
developed 
from multiple historical events
and movements
ranging from the Spanish Inquisition to social Darwinism, eugenics, and modern IQ tests, starting as early as the fifteenth century, when the Catholic Church in Spain persecuted the Jewish population for \say{impurity of blood} \cite{10.2307/j.ctt9qdt73.5}. 

He
goes on to explain that some Enlightenment scholars like David Hume and Immanuel Kant believed that, based on skin colours, there are more than one race of humans, and that 
white men 
are
the most civilized
people \cite{10.2307/j.ctt9qdt73.8}. In the nineteenth century, drawing from 
evolution theory, social Darwinists like Herbert Spencer argued that helping the poor and the weak was an interference with natural selection, coining the term \say{survival of the fittest}. 
This
led to sterilization and ultimately the extermination camps of the eugenics movement \cite{10.2307/j.ctt9qdt73.8}. 

Moving to the 1970s, 
\citet{10.2307/j.ctt9qdt73.13} shows that 
Arthur Jensen, a professor of Educational Psychology at the University of California, argued that 
Black people 
are intellectually 
inferior to 
white people. This argument was reasserted in the 1990s with the publication of Richard Herrnstein and Charles Murray's 
\textit{The Bell Curve}.

\citet{10.2307/j.ctt9qdt73.14}
goes on to show that in the 2000s, racism took
on
a 
disguise 
of
\say{Culturism}, 
as coined by the anthropologist Franz-Boas to explain the difference in human behaviour and social organizations. Culturism paved the way to modern-day anti-immigration agendas with immigrants, like Arabs or Muslims, not claimed to be genetically inferior to Europeans, but to have a cultural burden that prevents them from integrating in 
the West.

\textbf{Homophobia} is intertwined with racism, as argued by 
\citet{Morris2010HistoryOL} in their research on the history of the LGBTQ community social movement.
Morris explains that homosexuality and transgender 
identity
were accepted in many ancient societies like those of ancient Greece, Native Americans, North Africa, and the Pacific Islands. These accepting cultures oppose the Western culture of heterosexuality and binary genders, who regarded homosexuality and transgender as foreign, savage, and evidence of inferior races. When Europeans started colonization campaigns, they imposed their moral codes and persecuted LGBTQ communities. The first known case of punishing homosexuality by death was in North America in 1566. 
Later, in 
the era of sexology studies in 1882 and 1897,  
European doctors and scientists labelled homosexuality as degenerate and abnormal,
and as recently as 
the 1980s and 1990s, 
AIDS was widely rationalised as being god's punishment for gay people. 

As argued by \citet{perez2019invisible} in
\textit{Invisible Women: Data Bias in a World Designed for Men},
\textbf{Sexism} can be tracked back 
to the fourth century B.C. when Aristotle articulated that the male form is the default form as an inarguable fact. 
This concept still persists today, as we can see in the one-size-fits-men approach to designing supposedly gender-neutral products like piano keyboards and smartphones.
Overall, as \citet{manne-2018-down} describes it in \emph{Down Girl: The Logic of Misogyny}, sexism consists of \say{assumptions, beliefs, theories, stereotypes, and broader cultural narratives that \ldots make rational people more inclined to support and participate in patriarchal social arrangements}.

\textbf{Marginalization} has been studied in social sciences by many scholars in critical race theory \cite{benjamin2019race}, gender studies \cite{mcintosh2001white, davis1982women}, and LGBTQ studies \cite{fausto2008myths}. However, negative stereotyping, stigma, and unintended bias continue against marginalised people based on ethnicity, religion, disability, sexual orientation, or gender. These stigmas and unintended bias have led to different forms of discrimination from education, job opportunities, health case, housing, incarceration, and others, as 
\citet{Nordell2021}
details in 
\textit{The End of Bias}. 


They can also have negative impact on the cognitive ability, 
and
mental and physical health of the people who carry 
their load.
As 
\citet{steele2011whistling} shows in
\textit{Whistling Vivaldi: How Stereotypes Affect Us and What We Can Do}, based on experiments 
in behavioural psychology,
carrying stigma made women underperform in maths tests, and African-American students underperform in academia.
Hence, stereotypes become 
self-fulfilling prophecies, 
eventually leading to 
their perpetuation 
and the continuation of the prejudice and discrimination.

In the age of knowledge, computing, and big data,
\textbf{prejudice and discrimination} have found their way to machine learning models. These models that are now dictating every aspect of our lives from 
online advertising, to employment and judicial systems that rely on black box models and discriminate against marginalised groups, while benefitting privileged elites, as 
\citet{o2017weapons}
explains in 
\textit{Weapons of Maths Destruction}.
One of the most 
well-known examples of discriminative decisions made by a machine learning models is the COMPAS 
algorithm, 
a risk assessment tool that measures the likelihood that a criminal becomes a recidivist, a term used in legal systems to describe a criminal who reoffends. 
Despite
Northpoint, the company that produced the COMPAS tool not sharing how the model measures the recidivism scores, the algorithm was deployed by the state of New York
in 2010. In 2016, ProPublica found that Black defendants are more likely than white defendants to be incorrectly judged to be at a higher risk of recidivism while 
the latter
were more likely than Black defendants to be incorrectly flagged as low risk \cite{Larson-et-al-2016-recidivism}. 

One example of algorithmic gender discrimination is the CV screening model
used by Amazon, which, according to a Reuters report in 2018, favoured CVs of male 
over female candidates even when both had the same skills and qualifications \cite{dastin-amazon-sexist-model}. Similar examples of algorithmic discrimination can be found against the LGBTQ community \cite{algorithms-homophobia}, older people \cite{algorithm-agism}, Muslims \cite{algorithm-islamophobia}, and people with disabilities \cite{algorith-disabled}.

\section{Bias and fairness: Definitions}
The term \textit{bias} is defined and used in many ways \citep{olteanu2019}. The normative definition of bias, 
in cognitive science is: ``behaving according to some cognitive priors and presumed realities that might not be true at all'' \citep{munoz2021}. 
The statistical definition of bias is ``systematic distortion in the sampled data that compromises its representatives'' \citep{olteanu2019}. 

In NLP, while bias and fairness have been described in several ways,
the statistical definition is most dominant \cite{elsafoury_sos_2022, Caliskan-etal-2017-weat, Garg2017, nangia-etal-2020-crows, nadeem-etal-2021-stereoset}. In the last two years or so, there has been a trend to distinguish two types of bias in NLP systems: intrinsic bias and extrinsic bias \cite{cao2022, kaneko2022, steed-etal-2022-upstream}. \textbf{Intrinsic bias} is used to describe the biased representations of pre-trained models.  
As far as we know,
there is no formal definition of intrinsic bias in the literature. However, from the research done to study bias in word embeddings \cite{elsafoury_sos_2022}, we can infer the following definition:  Intrinsic bias is \textit{stereotypical representations of certain groups of people learned during pre-training}. For example, when a model associates women with certain jobs like caregivers and men with doctors \cite{Caliskan-etal-2017-weat}. This type of bias exists in both static 
\cite{Caliskan-etal-2017-weat, Garg2017} and contextual word embeddings \cite{nangia-etal-2020-crows, nadeem-etal-2021-stereoset}.

On the other hand, \textbf{Extrinsic bias}, also known as model fairness, has many formal definitions 
built on those
from literature on the fairness of exam testing from the 1960s, 70s and 80s \cite{Hutchinson-and-Mitchel-2019-50-years}.  The most recent 
fairness definitions are broadly categorized into two groups: \textbf{Individual fairness}, which is defined as \say{\textit{An algorithm is fair if it gives similar predictions to similar individuals}} \cite{kusner2017counterfactual}. 

For a given model $\hat{Y}: X \rightarrow Y$ with features $X$, sensitive attributes $A$, prediction $\hat{Y}$, and two individuals $i$ and $j$, and if individuals i and j are similar. The model achieves individual fairness if 
\begin{equation}
    \hat{Y}(X^i, A^i) \approx \hat{Y}(X^j, A^j)
\end{equation}

The second type of fairness definition is \textbf{Group fairness}, which can be defined as \say{\textit{An algorithm is fair if the model prediction $\hat{Y}$ and sensitive attribute $A$ are independent}} \cite{caton2020fairness, kusner2017counterfactual}. Based on group fairness, the model is fair if 

\begin{equation}
    \hat{Y}(X| A=0) = \hat{Y}(X|A=1)
\end{equation}

Group fairness is the most common definition used in 
NLP. There are different ways to measure it, like equality of odds \cite{baldini-etal-2022-fairness}.
However, other
metrics 
have been
proposed in the NLP literature to measure individual fairness like counterfactual fairness methods \cite{prabhakaran-etal-2019-perturbation}.  

\begin{figure*}[ht!]
    \centering
\includegraphics[width=0.7\textwidth]{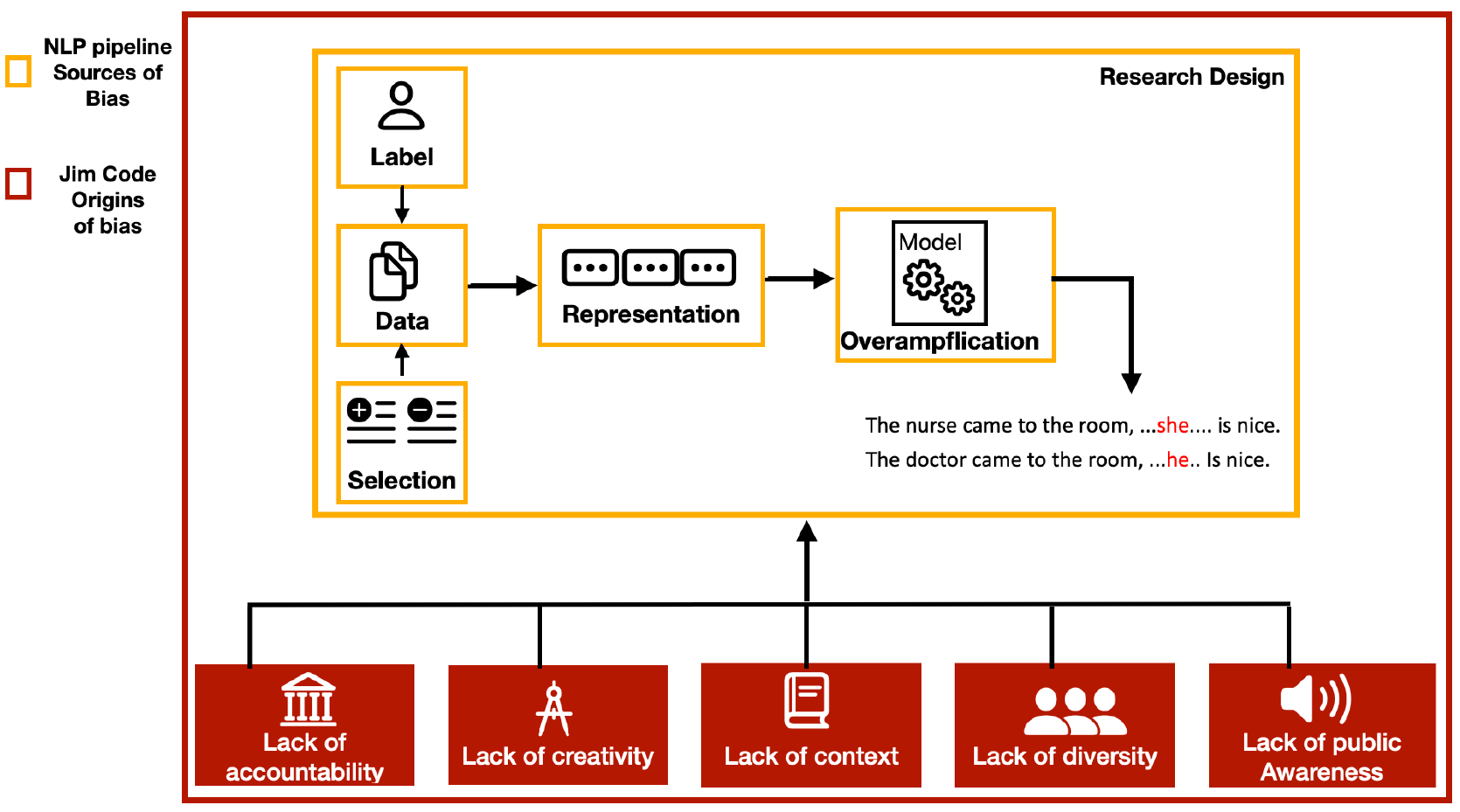}
    \caption{The origins of bias in supervised NLP models}
    \label{fig:sources_of_bias}
\end{figure*}

\section{Origins of bias}
While much 
literature 
proposes methods to measure bias and fairness in 
NLP models, there are far fewer papers that discuss their origins. 
Those that do so tend to neglect
literature from social science or the critical race theory that has examined topics directly related to bias like racism, sexism, or homophobia. This short-sightedness has, so far, led to cosmetic changes in the proposed NLP models to fix the problem of bias rather than fixing the racist, sexist, homophobic status quo \cite{benjamin2019race}. 

In this section, we review the different origins of bias in NLP systems from the Jim Code perspective of social science, using tools like critical race theory, digital studies, gender studies, LGBTQ studies, and internet and data activism. Then we review the sources of bias from a purely 
NLP perspective, while trying to connect these two 
strands
to 
gain 
a more profound understanding of the origins of bias. Figure \ref{fig:sources_of_bias} shows an overview summary of the origins of bias from the different perspectives, and how the biases in the NLP pipeline originate in the Jim Code.

\subsection{The Jim Code perspective}
\label{sec:jim_code}
As previously described, Jim Code is a term that refers to the new forms of systematic discrimination found in new technologies that build on older discriminatory systems. This is one of the main origins of bias and unfairness that we find in most NLP systems. This can be broken down into the following sources of bias:
\begin{enumerate}[wide=0pt]
    \item \textbf{Lack of context:} In 
    \textit{More than a Glitch}, 
    \citet{Broussard2023}
    explains that, like computers, the data used to train NLP and ML models are produced without a specific human context. A similar point is made by \citet{benjamin2019race}, who discusses 
    how 
    social and historical contexts are not taken into consideration when data is 
    collected to train NLP models. But it is not only the data. With the NLP models being developed in isolation from social science
    perspectives,
    how these systems impact the people of different identity groups gets overlooked. For example, models 
    output decisions on
    who is eligible to get a loan or get a job without consideration of the fact that this might increase the wealth gap between marginalised and privileged groups. 
    
    Moreover, it is because of the lack of context that researchers in NLP do not think about the harmful ways that their proposed 
    systems could be used. For example, when models are used to detect race from last names and zip codes, their developers have probably failed to consider how they will be employed by certain businesses 
    to illegally collect information on ethnicity \cite{benjamin2019race}. 
    Even greater harm is caused when
    a model categorises people as  criminals or terrorists due to their inferred ethnicity.

    \item \textbf{Lack of creativity:} Because of the lack of context, many developers of ML and NLP models tend to build their systems on top of existing racist, sexist, homophobic, ageist, and ableist systems. 
    An example is when recommender systems used \say{cultural segregation} to infer information about a person's ethnicity to personalise their recommendations, using ethnicity as a proxy for individuality \cite{benjamin2019race}. Hence, those systems perpetuate the racist view that 
    people who belong to a specific group must have similar preferences. Researchers need to be more creative and find other ways to recommend content that do not rely on social bias shortcuts.

    \item \textbf{Lack of accountability:} There is a lack of accountability that allows tech companies to maximize profits and get away with creating oppressive systems that are not just \say{glitches} as explained by critical race and digital humanities studies activists~\citep{Broussard2023,nobel2018,benjamin2019race}.
    A lack of accountability enables companies to sell their systems as black boxes without explaining how their models make decisions \cite{o2017weapons}. We also see that in the scientific community, where big tech companies publish papers 
    emphasising
    their models' excellent results without sharing those models or the data that were used to train them, precluding reproducibility.
    
    Moreover, when, the Justice League, a group of AI ethicists and activists,  launched the Safe Face pledge to ensure that computer vision models don't discriminate between people based on their skin colour, no major tech company was willing to sign it \cite{benjamin2019race}. With the lack of accountability and legislation, big tech companies, which are one of the main drivers of the field, have no reason to revise and change the way they build their ML and NLP systems, or to include the social and historical context into their research in a way that profoundly changes the systems instead of just covering it up and fixing the \say{glitches}.
    
    \item \textbf{Lack of diversity:} The majority of ML and NLP technologies are developed in companies or research institutes in Western societies and by researchers who are mostly white, able-bodied, heterosexual men. They develop and test 
    systems that work well for them, without considering how functional these systems are for people from different backgrounds. Examples are facial recognition systems that only work with people with light skin \cite{benjamin2019race, Broussard2023} and  CV recommendation systems that favour applicants with male names \cite{dastin-amazon-sexist-model}. There is also a lack of diversity when it comes to the targeted customers of the 
    systems. Since most of these technologies are expensive to buy, the developers of these systems focus on the customers who can afford it and who are also predominantly white, able-bodied, heterosexual men \cite{benjamin2019race}. This lack of diversity, in addition to the lack of social and historical contexts, leads to the development of discriminatory systems.
    
    \item \textbf{Lack of public awareness:}
    In addition to the previously discussed origins of bias in NLP, another factor that allows the biases 
    to spread is the lack of public awareness. This 
    is a result of using mathematical and statistical terminology and jargon that most non-specialists can't understand. This lack of understanding of how ML and NLP models work and their limitations led people to over-trust AI systems and leads to \say{Technochauvinism}, described by 
    \citet{Broussard2023}
    as:
    
    \begin{quote}
    \textit{\say{the kind of bias that considers computational solutions to be superior to all other solutions. Embedded in this bias is a priori assumption that computers are better than humans which is actually a claim that the people who make and program computers are better than other humans.}}
    \end{quote}

    The lack of public awareness and Technochauvinism are the reasons why banks, schools, hospitals, universities, and other institutions that are supposed to deal with people and society and make social decisions adopting  NLP systems that are poorly understood, with the false notion that they are unbiased, and their decisions are faultless and objective \cite{benjamin2019race,Broussard2023}. 
\end{enumerate}

\subsection{The NLP pipeline perspective}
\label{sec:nlp_perspective}
We now turn to the sources of bias in the NLP pipeline described in the literature. \citet{shah-etal-2020-predictive} introduce four sources of bias in the NLP pipeline that might impact the model's fairness. \citet{hovy2021five} also discuss these, adding a fifth source related to the overarching design of NLP research projects.

Here, we outline these pipeline sources of bias and also show how they, in fact, originate in the Jim Code perspective.

\begin{enumerate}[wide=0pt]
    \item \textbf{Research design:}
    According to \citet{hovy2021five}, research design bias is manifested in the skewness of NLP research towards Indo-European languages, especially English.
    This skew leads to a self-fulfilling prophecy, since  most of the research focuses on text in English, more data in English becomes available, which in turn makes it easier for NLP researchers to work on English text. 
    This has further ramifications as \citet{hovy2021five} also question whether, if English was not the \say{default} language, the $n-gram$ would have been the focus of NLP models. The authors argue that the \emph{lack of diversity} in the makeup of NLP research groups, is one of the reasons behind the linguistic and cultural skewness in NLP research.

    In addition to these skews,
    there are further sources of bias reflected in research design that originate from the Jim Code perspective. 
    \emph{Lack of social context} is clearly manifested in NLP research design. For example, NLP researchers deal with language as a number of word occurrences and co-occurence probabilities rather than dealing with language as a diverse social component that reflects societal relationships and biases \cite{holmes2013introduction}. Another example, is \emph{lack of historical context}, with most of the data that NLP models are trained on  generated by white middle-class men, resulting in speech recognition models not recognizing African American dialects \cite{benjamin2019race,tatman-2017-gender} and hate speech detection models falsely flagging African American dialect as hateful \cite{sap2019}.  \emph{Lack of creativity} is also reflected in research design. For example, with NLP models relying on the $n-gram$ models and words co-occurrences, they incorporate biases such that they associate gendered words,\say{woman} and \say{man}, with certain jobs, \say{nurse} and \say{doctor} \cite{Caliskan-etal-2017-weat}. As \citet{hovy2021five} contend, \emph{lack of diversity} is also reflected in the research design bias, as evident in the skewness towards Indo-European languages. Because of the \emph{lack of accountability} and the \emph{lack of public awareness}, NLP research design bias has been going on for decades, largely unnoticed and unconsidered.

    \item \textbf{Selection bias:} 
    Selection bias is a result of non-representative observations in the datasets used to train NLP models \cite{shah-etal-2020-predictive, hovy2021five}. This bias could manifest when a model is trained on text data that has been generated by one group of people, but is subsequently deployed in the real world and used by more diverse groups.
    For example, the syntactic parsers and part-of-speech taggers that were trained on data generated by white middle-aged men, which then impacted the accuracy of these models when tested on text generated by different groups of people \cite{shah-etal-2020-predictive}. Another example in hate speech detection models, where the models were trained on data with over-representation of terms associated with marginalised identity groups with the positive class (hateful) resulting in the models falsely labelling content as hateful just because it includes mentions of those identities \cite{sap2019, Dixon2018}.

    Selection bias is also a result of \emph{lack of context}, since the NLP researchers used datasets with over-representation of one group and under-representation of many other groups due to their lack of social and historical context of who generated that data and which identity groups are under-represented in the chosen data. \emph{Lack of diversity} is also a prominent reason behind selection bias in NLP, as most of the researchers come from non-marginalised backgrounds \cite{michael2022nlp} with blind spots for the under-represented groups of people. Finally, \emph{lack of creativity} is another reason behind selection bias. As NLP researchers build their models on biased systems that generated biased data, instead of being  more creative and using more diverse representative data that work for everyone.

    \item \textbf{Label bias:} Label bias, also known as annotator bias, is a result of a mismatch between the annotators and the authors of the data. There are many reasons behind label bias. It can result from spamming annotators who are uninterested in the task and assign labels randomly to get the task done, as can happen on crowdsourcing platforms. It also happens due to confusion or ill-designed annotation tasks. Another reason is due to the individual annotator's perception and interpretation of the task or the label \cite{hovy2021five}. Moreover, there could be a mismatch between the authors' and annotators' linguistic and social norms. For example, annotators are prone to mislabel content as hateful for 
    including
    the N-word, 
    despite its often benign in-group use by African Americans.
    Finally, 
    labels might carry 
    the annotators
    societal perspectives and social biases \cite{sap2019}.

    On the other hand, we can argue that some of these biases result from 
    unfairness in the crowdsourcing systems. Since the pay that annotators receive is 
    often extremely low
    they are incentivised to complete as many tasks as they can as fast as possible to make ends meet, which in turn impacts the quality of the labels \cite{fort-etal-2011-last}. Moreover, \citet{10.1145/3492853} argue that the bias in the labels is not only due to the biased perceptions of the annotators, but also due to a certain format the annotators have to follow for their annotation tasks and if that format falls short on diversity, the annotators lack the means to communicate that to the designers of the task. 
    An
    example is 
    when an annotator is 
    presented with a binary gender choice
    even if the data contains information about non-binary or transgender people. Hence, label bias could be seen as a result of the \emph{lack of context}. As the NLP researchers who mismatch the demographics of their data's authors and annotators do that due to lack of social context of the author of the data. Label bias is also a result of the \emph{lack of accountability}, as big tech and NLP research groups hire annotators with unfair pay in addition to the lack of means for those annotators to communicate problems in the annotation task with the task designer due to power dynamics.

    \item \textbf{Representation bias:} 
    Representation bias, also known as intrinsic bias or semantic bias, describes the societal stereotypes that language models encode during pre-training. The bias exists in the training dataset that then gets encoded in the language models static \cite{Caliskan-etal-2017-weat, elsafoury_sos_2022, Garg2017}, or contextual \cite{nangia-etal-2020-crows, nadeem-etal-2021-stereoset}. \citet{hovy2021five} argue that one of the main reasons behind representation bias is the objective function that trains the language models. As these objective functions aim to predict the most probable next term given the previous context, which in turn makes these models reflect our biased societies in the data.

    Again, representation bias is a result of the \emph{lack of social and historical context}, which is why NLP researchers tend to use biased data to train these language models. It is also a result of \emph{lack of creativity} as instead of using objective function that aim to reproduce the biased word that we live in, NLP researchers could have used different objective functions that optimize fairness and equality in addition to performance.
    
    \item \textbf{Model overampflication bias:} 
    According to \citet{shah-etal-2020-predictive}, overampflication bias happens because, during training, the models rely on small differences between sensitive attributes regarding an objective function and amplify these differences to be more pronounced in the predicted outcome. For example, in the imSitu image captioning dataset, 58\% of the captions involving a person in a kitchen mention women, resulting in models trained on such data predicting people depicted in kitchens to be women 63\% of the time \cite{shah-etal-2020-predictive}. For the task of hate speech detection, overampflication bias could happen because certain identity groups could exist within different semantic contexts, for example, when an identity group like \say{Muslims} co-occurs with the word \say{terrorism}. Even if the sentence does not contain any hate, e.g. \say{Anyone could be a terrorist not just muslims}, the model will learn to pick this information up about Muslims and amplify them, leading to these models predicting future sentences that contain the word ``Muslim'' as hateful.  According to \cite{hovy2021five}, one of the sources of overampflication bias is the choice of objective function used in training the model. Since these objective functions mainly aim to improve precision, the models tend to exploit spurious correlations or statistical irregularities in the data to achieve high performance by that metric. 

    Overamplification bias is again a result of the \emph{lack of social and historical context}, which results in using data that has an over-representation of certain identities in a certain social or semantic context. These over-representations are then picked up by the models during training. Another reason is the \emph{lack of creativity} that results in choosing objective functions that exacerbate the differences found in the datasets between different identity groups and prioritising overall performance over fairness.
\end{enumerate}

\section{Discussion}
It is clear that the sources of bias that we find in the NLP pipeline do not come out of nowhere, but have their origins in those that have been
outlined in
the social science, critical race theory and digital humanities studies---the Jim Code perspective.
Despite this, the bias metrics that have been proposed in the NLP literature measure only pipeline bias, which
has
led to limitations in the currently proposed methods to measure and mitigate bias. 

In this section, we outline these limitations and recommend measures to mitigate them.

\subsection{Limitations of bias research in NLP}
\label{sec:limitations}
The lack of 
scrutiny of
the social background behind biases, has led approaches to bias 
measurement to incorporate the same methods that introduced bias in the first place.
For example, crowdsourcing the data used in measuring bias in language models \cite{nangia-etal-2020-crows, nadeem-etal-2021-stereoset} reintroduces label bias into the metric that is supposed to measure bias. Moreover, studies that propose bias metrics in NLP don't incorporate the social science literature on bias and fairness, which results in a lack of articulation of what these metrics actually measure, and ambiguities and unstated assumptions, as discussed in \cite{Blodgett-etal-2021-norweigan-salmon}. 

This results in 
limitations to the current bias metrics proposed and used in the NLP literature. One of these is that different bias metrics 
produce
different bias scores, which makes it difficult to come to any conclusion on how biased the different NLP models are \cite{elsafoury-etal-2022-comparative}. There is also the limitation that current bias metrics claim to measure the existence of bias and not its absence, meaning that lower bias scores do not necessarily mean the absence of bias \cite{may-etal-2019-seat}, leading to lack of conclusive information about the NLP models. Another consequence of the lack of understanding what the bias metrics in NLP actually measure, is that most of the research done on investigating the impact of social bias in NLP models on the downstream tasks could not find an impact on the performance of the downstream tasks \cite{goldfarb-tarrant-etal-2021-intrinsic, elsafoury_sos_2022} or the fairness of the downstream tasks \cite{kaneko2022, cao2022}.

Similarly, one of the main limitations of the proposed methods to measure individual fairness metrics is that 
the motivation behind the proposed metrics and what the metrics actually measure are not disclosed. For example, \citet{prabhakaran-etal-2019-perturbation, czarnowska-etal-2021-quantifying, qian-etal-2022-perturbation} propose metrics to measure individual fairness using counterfactuals without explaining the intuition behind their proposed methods and how these metrics meet the criteria for individual fairness.  

As for group fairness metrics, they are all based on statistical measures that have come in for criticism. For example, \citet{hedden2021statistical}
argues that group fairness metrics are based on criteria that cannot be satisfied unless the models make perfect predictions or that the base rates are equal across all the identity groups in the datase. 
Base rate here refers to the class of probability that is unconditioned on the featural evidence \cite{BARHILLEL1980211}. \citet{hedden2021statistical} goes on to ask if the statistical criteria of fairness cannot be jointly satisfied except in marginal cases, which criteria then are conditions of fairness. 

In the same direction of questioning the whole notion of using statistical methods to measure fairness,
\citet{Broussard2023} argues that some of the founders of the field of statistics were white supremacists, which resulted in skewed statistical methods and suggests that to measure fairness, maybe we should use non-statistical methods.
Approaching the bias and fairness problem in  NLP as a purely quantitative problem led the community to develop quantitative methods to remove the bias from NLP models like \cite{Bolukbasi2016,liang2020,schick2021self} which resulted in only a superficial fix of the problem while the models are still biased \cite{kaneko2022,gonen-goldberg-2019-lipstick}. As shown above, Section \ref{sec:nlp_perspective}, bias and fairness in NLP models are the results of deeper sources of bias, and removing the NLP pipeline sources of bias would not lead to any real change unless the more profound issues from the social science perspective are addressed. 

Similarly, current efforts to improve the model's fairness have relied on quantitative fairness measures that aim to achieve 
equality
between different identity groups, when 
equality does not necessarily mean 
equity \cite{Broussard2023}. 
Achieving equality would mean that the NLP models give similar performances to different groups of people. However, in some cases, fairness or equity would require treating people of certain backgrounds differently.  For example, \citet{DiasOliva2021} demonstrate that Facebook's hate speech detection models restrict the use of certain words considered offensive without taking into consideration the context in which they are being used. This leads to the censoring of some of the comments written by members of the LGBTQ community, who claim some of these restricted words as self-expression. In this case, 
equality did not lead to 
equity.

\subsection{How to mitigate those limitations?}
\label{sec:recommendations}
Addressing the Jim Code sources of bias, is not a simple task. However, by doing so, we can take steps towards developing more effective ways to make  NLP systems more inclusive, fairer and safer for everyone. 
Here, we outline actionable recommendations for the NLP community:
\begin{enumerate}[wide=0pt]
\item \textbf{Lack of context} can be addressed by incorporating social sciences as part of the effort of mitigating bias in NLP models. This is only possible through:
\begin{enumerate}
    \item \emph{Interdisciplinary research} where scientists with backgrounds in fields such as critical race theory, gender studies and digital humanities studies are 
included in NLP project teams, so they can point out the social impact of the choices made by the NLP researchers.
\item  It can also be addressed by 
further \emph{integration of the teaching of data and machine learning ethics into NLP curricula},
whereby students gain an understanding of the societal implications of the choices they make. Currently, they are typically only exposed to minimal and tokenistic treatment of the topics of
bias and fairness in NLP models, which is insufficient to understand the origins of bias from a social science perspective. 
This should also include training in AI auditing, 
enabling students to assess the
limitations and societal impact of the NLP systems they develop \cite{Broussard2023}.
\end{enumerate}
 
\item \textbf{Lack of creativity} is a direct result of lack of context. We can address the lack of creativity by:
\begin{enumerate}
    \item Raising \emph{awareness of the social and historical context and the social impact of development choices} among NLP researchers. This will encourage more creative methods to achieve their goals, instead of the reproduction of oppressive systems in shiny new packaging. Online competition and code sharing platforms could be a place to start, for example, 
    creating shared tasks 
    in which participants develop new NLP models that do not rely on $n-grams$ or objective functions that do not amplify societal biases.

    \item Another way to encourage NLP researchers to re-investigate NLP fundamentals, is \emph{specialized conferences and workshops on re-imagining NLP models with an emphasis on fairness and impact on society}. This effort is already underway with conferences like ACM Conference on Fairness, Accountability, and Transparency (ACM FAccT) \footnote{\url{https://facctconference.org/}}. The outcomes of these endeavours should be open for auditing, evaluation and reproducibility. One way to achieve that, without the controversy of open-source, is for NLP conferences to adopt the ACM artifact evaluation measures\footnote{\url{https://www.acm.org/publications/policies/artifact-review-badging}} and give reproducibility badges to published papers. This could be developed further to give social responsibility badges to the papers that were audited by a special responsible NLP committee.

    \item \emph{Specialized interdisciplinary seminars} in major NLP conferences could encourage NLP researchers to collaborate with social scientists. For example, organizing events like the Dagstuhl seminars\footnote{\url{https://www.dagstuhl.de/en/seminars/dagstuhl-seminars}} that invite social scientists to discuss their work on bias and fairness which might lead to the exchange and development of ideas between social and NLP researchers.

\end{enumerate}

\item \textbf{Lack of diversity} can be addressed with:

\begin{enumerate}
    \item \emph{Greater diversity on research teams} working on NLP problems. A more diverse perspective will be introduced to the research to make sure that the proposed solution and new systems are inclusive and work for everyone. 
    
    \item \emph{NLP conferences play a great role in promoting diversity} in NLP research by incorporating shared tasks that encourage researchers to work on low-resourced languages. For example, the shared tasks\footnote{\url{http://nlpprogress.com/}} on Arabic, Persian, Korean, and others.

    \item \emph{Incorporating more diversity workshops} in NLP conferences that allow researchers from different backgrounds to publish their work, e.g. the WiNLP workshop\footnote{\url{https://www.winlp.org/}}. 
    
    \item This effort can go further by creating \emph{shared tasks that test the impact of NLP systems on different groups of people}.
\end{enumerate}

\item \textbf{Lack of accountability} The suggested measures should be enforced with:
\begin{enumerate}
    \item \emph{State level regulation} to make sure that research is not conducted in a way that may harm society, which is only possible by holding universities and big tech companies accountable for the systems they produce. One step taken in this direction is the EU AI Act\footnote{\url{https://artificialintelligenceact.eu/}} which is a legislative proposal that assigns AI applications to three risk categories: 
\begin{quote}   
\textit{\say{First, applications and systems that create an \textbf{unacceptable risk}, such as government-run social scoring of the type used in China, are banned. Second, \textbf{high-risk applications}, such as a CV-scanning tool that ranks job applicants, are subject to specific legal requirements. Lastly, applications not explicitly banned or listed as high-risk are largely left unregulated.}}
\end{quote}

\item There should also be an \emph{AI regulation team} that works for governments and employs AI auditing teams and social scientists to approve newly developed NLP systems before they are released to the public.  \cite{Broussard2023}
\end{enumerate}

\item \textbf{Lack of awareness and Technochauvinism} The suggested regulations, can only happen by democratically electing people who are willing to put these regulations in place. This comes with raising the awareness of the limitations of the current ML and NLP systems. It is important that the public is aware that the likely doomsday scenario is not an AI system that outsmarts humans and controls them, but one that behaves like a Stochastic Parrot \cite{gebru-stocatic-parrot-2021} that keeps reproducing our discriminative systems on a wider scale under the mask of objectivity \cite{o2017weapons, benjamin2019race, Broussard2023, nobel2018}. NLP researchers can help to raise public awareness through:

\begin{enumerate}
    \item \emph{Journalism} is an important resource to inform the public of the limitations and ethical issues in the current AI systems. Muckraking journalists in ProPublica, and The New York Times investigate AI technologies, sharing their investigations with the public \cite{Broussard2023}. For example, the journalist's investigation of the COMPAS system and its unfairness was published by ProPublica. NLP researchers should be encouraged to accept interview invitations from journalists to share their worries about the limitations of the current NLP systems.

    \item \emph{Published Books for non-specialists} is another way to raise public awareness on issues related to discrimination in AI systems. Especially books that are targeted at non-specialists. For example, books like \textit{Race after Technology, More than a Glitch, and Algorithms of Oppression}. NLP researchers can participate in those efforts by writing about current NLP systems and their limitations for non-specialists. 

    \item \emph{Talks}: NLP researchers should be encouraged to share their views on AI in non-academic venues. For example, participating in documentaries like \textit{Coded Bias}\footnote{\url{https://www.imdb.com/title/tt11394170/}} could bring awareness to the public.  

    \item \emph{Museums} of technology, and arts could also raise public awareness of the limitations and potential dangers of AI. For example, in 2022, the Modern Museum of Arts, had an exhibition called \say{Systems}\footnote{\url{https://www.moma.org/collection/works/401279?sov_referrer=theme&theme_id=5472}}, showing how AI systems work, their inequalities, and how much natural resources are used to build them. NLP researchers and universities can help organize exhibitions on the limitations of NLP systems.
    
    \item \textit{Social media awareness campaigns} could be a way to reach more people, especially younger people. Currently, individual NLP researchers share on social media their views and worries about NLP systems. However, an organized campaign can be more effective.
\end{enumerate}
\end{enumerate}

\section{Conclusion}
In this paper, we have reviewed the literature on historic forms of sexism, racism, and other types of discrimination that are being reproduced in the new age of technology on a larger scale and under the cover of supposed objectivity in NLP models.  We reviewed the origins of bias from the NLP literature in addition to the social science, critical race theory, and digital humanities studies literature. We argue that the sources of bias in NLP originate in those identified in the social sciences, and that they are direct results of the sources of bias from the \say{Jim Code} perspective. We also demonstrate that neglecting the social science literature in attempting to build unbiased and fair NLP models has led to unreliable bias metrics and ineffective debiasing methods. We argue that the way forward 
is to incorporate knowledge from social sciences and further collaborate with social scientists to make sure that these goals are achieved effectively without negative impacts on society and its diverse groups. Finally, we share a list of actionable suggestions and recommendations with the NLP community on how to mitigate the discussed Jim code origins of bias in NLP research.

\section{Ethical Statement}
Our critical review on the origins of bias in NLP systems should not produce any direct negative impacts or harms. However, our work does not come without risks. 
One of these could be in discouraging quantitative research on bias and fairness in NLP by making such work seem daunting, requiring collaborations and more effort than other research disciplines in NLP. 
However, our aim is 
rather to encourage researchers to be more cautious and take a more inclusive approach to their research, incorporating social scientists and their knowledge into efforts at understanding bias in NLP.

\bibliographystyle{acl_natbib}
\bibliography{main.bib}
\end{document}